\documentclass{llncs}
\usepackage{array,multirow,graphicx}
\usepackage[english]{babel}
\usepackage{cite}
\usepackage{lipsum}
\usepackage{graphicx}
\usepackage{amssymb}
\usepackage{enumitem}
\usepackage{xfrac}
\usepackage{amsmath}
\usepackage{url}
\usepackage{url}
\usepackage{lineno,hyperref}
\usepackage{array}
\usepackage[utf8]{inputenc}
\usepackage{csquotes}
\hypersetup{
    colorlinks=true,
    linkcolor=blue,
    urlcolor=blue,
		citecolor=blue,
}
\usepackage{subcaption}
\usepackage{booktabs}

\begin{document}
\title{A Scheme for Continuous Input to the Tsetlin Machine with Applications to Forecasting Disease Outbreaks \vspace{-3mm}}

\author{K. Darshana Abeyrathna \and
Ole-Christoffer Granmo \and
Xuan Zhang \and Morten~Goodwin}
\institute{Centre for Artificial Intelligence Research, University of Agder, Grimstad, Norway
\email{darshana.abeyrathna@uia.no, ole.granmo@uia.no, xuan.z.jiao@gmail.com, morten.goodwin@uia.no}}
\maketitle 
\begin{abstract}
\vspace{-6mm}
In this paper, we apply a new promising tool for pattern classification, namely, the \emph{Tsetlin Machine} (TM), to the field of disease forecasting. The TM is interpretable because it is based on manipulating expressions in propositional logic, leveraging a large team of Tsetlin Automata (TA). Apart from being interpretable, this approach is attractive due to its low computational cost and its capacity to handle noise. To attack the problem of forecasting, we introduce a preprocessing method that extends the TM so that it can handle continuous input. Briefly stated, we convert continuous input into a binary representation based on thresholding. The resulting extended TM is evaluated and analyzed using an artificial dataset. The TM is further applied to forecast dengue outbreaks of all the seventeen regions in Philippines using the spatio-temporal properties of the data. Experimental results show that dengue outbreak forecasts made by the TM are more accurate than those obtained by a Support Vector Machine (SVM), Decision Trees (DTs), and several multi-layered Artificial Neural Networks (ANNs), both in terms of forecasting precision and F1-score.     

\keywords{Tsetlin Machine  \and Tsetlin Automata \and Learning Automata \and Pattern Recognition with Propositional Logic \and Disease Outbreaks Forecasting.}
\end{abstract}
\vspace{-9mm}
\section{Introduction}
\vspace{-1mm}

\begin{figure}[t]
\vspace{-4mm}
\centering
\includegraphics[width=8.5cm]{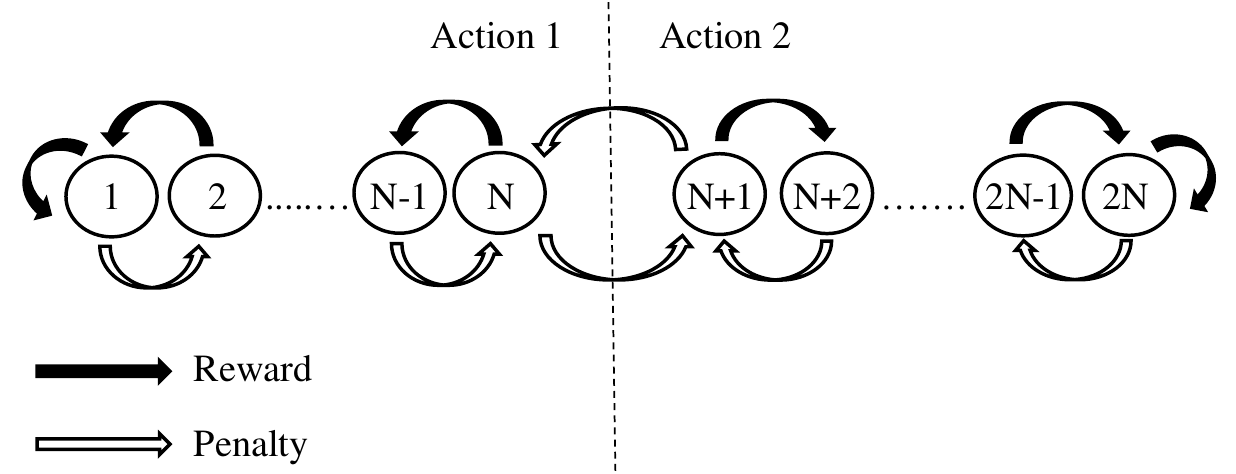}
\caption{Transition graph of a two-action Tsetlin Automaton.} \label{fig1}
\vspace{-7mm}
\end{figure}

The Tsetlin Machine (TM) is a recent pattern classification method that manipulates expressions in propositional logic based on a team of Tsetlin Automata (TAs) \cite{Ole1}. A Tsetlin Automaton (TA) is a fixed structure deterministic automaton that learns the optimal action among the set of actions offered by an environment. Fig. \ref{fig1} shows a two-action TA with 2N states. The action that the TA performs next is decided by the present state of the TA. States from 1 to N maps to Action 1, while states from N+1 to 2N maps to Action 2. The TA interacts with its environment in an iterative way. In each iteration, the TA performs the action associated with its current state. This, in turn, randomly triggers a reward or a penalty from the environment, according to an unknown probability distribution. If the TA receives a reward, it reinforces the action performed by moving to a “deeper” state, one step closer to one of the ends (left or right side). If the action results in a penalty, the TA moves one step towards the middle state, to weaken the performed action, ultimately jumping to the middle state of the other action. In this manner, with a sufficient number of states, a TA converges to performing the action with the highest probability of producing rewards -- the optimal action -- with probability arbitrarily close to unity, merely by interacting with the environment\cite{narendra3}.

The TM, introduced in 2018 by Granmo \cite{Ole1}, uses the TA as a building block to solve complex pattern recognition tasks. The TM operates as follows. Firstly, propositional formulas in disjunctive normal form are used to represent patterns. The TM is thus a general function approximator. The propositional formulas are learned through training on labelled data by employing a collective of TAs organized in a game. The payoff matrix of the game has been designed so that the Nash equilibria (NE) correspond to the optimal configurations of the TM. As a result, the architecture of the TM is relatively simple, facilitating transparency and interpretation of both learning and classification. Additionally, the TM is designed for bit-wise operation. That is, it takes bits as input and uses fast bit manipulation operators for both learning and classification. This gives the TM an inherent computational advantage.  Experimental results show that TM outperforms ANNs, Support Vector Machines (SVMs), the Naïve Bayes Classifier (NBC), Random Forests (RF), and Logistic Regression (LR) in diverse benchmarks \cite{Ole1,Berge2018}. These promising properties and results make the TM an interesting target for further research.

In this paper, we introduce a novel scheme that improves the accuracy of the TM when features are continuous. In Section~2, we provide an overview of related work. Then, in Section~3, we present our scheme for handling continuous features. In all brevity, we encode  continuous features in binary form based on thresholding. The behavior of the resulting TM is studied in Section~4 based on both an artificial dataset and real-life data, focusing on dengue fever forecasting. Section~5 summarizes our research and provides pointers for further work.

\vspace{-2mm}
\section{Related Work}
\vspace{-1mm}

Propositional logic is a well-explored framework for knowledge based pattern classification. In \cite{ogihara7}, Disjunctive Normal Form (DNF) is used to represent the patterns in clinical and genomic data to find the recurrence of liver cancer. Data is converted to bits by setting thresholds for continuous features. Based on the input features, logical functions for recurrence and non-recurrence are created. Another example is the use of Boolean expressions to capture visual primitives for visual recognition, rather than relying on a data driven approach \cite{cruz8}. In all brevity, the advantage of propositional logic for pattern classification, and knowledge based approaches in general, as opposed to data driven statistical models, is that patterns can be identified even without a single training sample.

Learning propositional formulas to represent patterns in data has a long history \cite{wang6}. Feldman investigates the hardness of learning DNF \cite{feldman9}, Klivans use Polynomial Threshold Functions to build logical expressions \cite{klivans10}, while Feldman leverages Fourier analysis\cite{feldman11}. Furthermore, so-called Probably Approximately Correct (PAC) learning has provided fundamental insight into machine learning, as well as providing a framework for learning formulas in DNF \cite{valiant12}. An integer programming approach is applied in \cite{hauser13} to learn disjunctions of conjunctions, providing promising results based on a Bayesian method. In addition to the above techniques, association rule mining models have been extensively applied in \cite{rudin14,mccormick15} to predict sequential events using set of rules. Recent approaches combine Bayesian reasoning with propositional formulas in DNF for robust learning of formulas from data \cite{wang6}. However, these techniques still suffer when facing noisy non-linear data, which may trap the learning mechanisms in local optima.

An attractive property of TA is that they support online learning in particularly noisy environment. Over several decades the basic TA, shown in Fig. \ref{fig1}, has been extended in several directions. These extensions include the Hierarchy of Twofold Resource Allocation Automata (H-TRAA) for resource allocation \cite{granmo16} and the stochastic searching on the line algorithm by Oommen et al. \cite{oommen19}. Furthermore, teams of Tsetlin Automata have been used to create a distributed coordination system \cite{tung17}, to solve the graph coloring problem \cite{bouhmala18}, and to forecast dengue outbreaks in the Philippines \cite{darshana20}. The TM is a recent addition to the field of TA, addressing complex pattern recognition.

In order to attack the problem of forecasting, this paper introduces a preprocessing method that extends the TM so that it can handle continuous input. To achieve this, we convert continuous input into a binary representation based on thresholding. We use an artificial dataset as well as a real-life dataset to evaluate this approach, namely, forecasting of dengue fever outbreaks in the Philippines. 

Different techniques have already been applied to forecast dengue outbreaks in different regions of the world. For instance, Seasonal Autoregressive Integrated Moving Average model is applied to forecast future dengue incidences in Guadeloupe \cite{gharbi21} and Bangladesh \cite{choudhury22}. In their research, temperature is identified as the best weather parameter to improve the forecasting performances. Here, 1-month ahead forecasting produces the highest accuracy, compared to 3-months and 1-year ahead forecasting. Similarly, dengue incidences in Rio de Janeiro, Brazil \cite{luz23}, Northeastern Thailand \cite{silawan24}, and Southern Thailand \cite{promprou25} are forecasted using Auto-regressive Integrated Moving Average models. Phung et al. investigate the forecasting ability of three regression models on dengue fever incidences in Can Tho city in Vietnam \cite{phung26}. They find that a Standard Multiple Regression model provides poor forecasting capability. However, the Poisson Distributed Lag model performs well in 12-months ahead forecasting and Seasonal Autoregressive Integrated Moving Average model performs well in 3-months ahead forecasting. The importance of utilizing data from neighboring regions to forecast dengue incidences is identified in \cite{abeyrathna27}, using an Artificial Neural Network as the forecasting model with data from the Philippines.

In contrast to the above approaches, we will here investigate whether a rule based approach, based on the Tsetlin Machine, can forecast outbreaks surpassing a decision threshold, across the regions of the Philippines.

\vspace{-4mm}
\section{Methodology}
\vspace{-1mm}

\subsection{The Tsetlin Machine Architecture}

The TM addresses pattern classification problems where a class can be represented by a collection of sub-patterns, each fixing certain features to distinct values. The TM is designed to uncover these sub-patterns in an effective, yet relatively simple manner.  In all brevity, the TM represents a class using a series of clauses. Each clause, in turn, captures a sub-pattern by means of a conjunction of literals, where a literal is a propositional variable or its negation. Each propositional variable takes the value \textit{False} or \textit{True} (in bit form, $0$ or $1$ respectively).

 Let $\textbf{\textit{X}} = [x_1, x_2, x_3, \ldots, x_n]$ be a feature vector consisting of $n$ propositional variables $x_k$ with domain $\{0,1\}$. Now suppose the pattern classification problem involves \textit{q} outputs, and \textit{m} sub-patterns per output that we need to recognize. Then the resulting pattern classification problem can be captured using $q \times m$ conjunctive clauses $C_i^j$,  $1 \leq j \leq q$, 1 $\leq$ \textit{i} $\leq$ \textit{m}. The output $y^j$, 1 $\leq$ \textit{j} $\leq$ \textit{q}, of the classifier is given as:
\begin{equation}\label{Eq1}
C_i^j = 1 \land \left(\bigwedge_{k \in I_i^j} x_k\right) \land \left( \bigwedge_{k \in \bar I_i^j} \lnot x_k\right).
\end{equation}

\begin{equation}\label{Eq2}
y^j = {\bigvee}_{i=1}^{m} C_i^j.
\vspace{2mm}
\end{equation}
Above, $I_i^j$ and $\bar I_i^j$ are non-overlapping subsets of the input variable indexes, $I_i^j, \bar I_i^j \subseteq \{1,.....n\},$ $I_i^j \cap \bar I_i^j = \emptyset$. The subsets decide which of the propositional variables take part in the clause, and whether they are negated or not.

\begin{figure}[t]
\vspace{-5mm}
\begin{subfigure}{1\linewidth}
\centering
\includegraphics[scale=0.64]{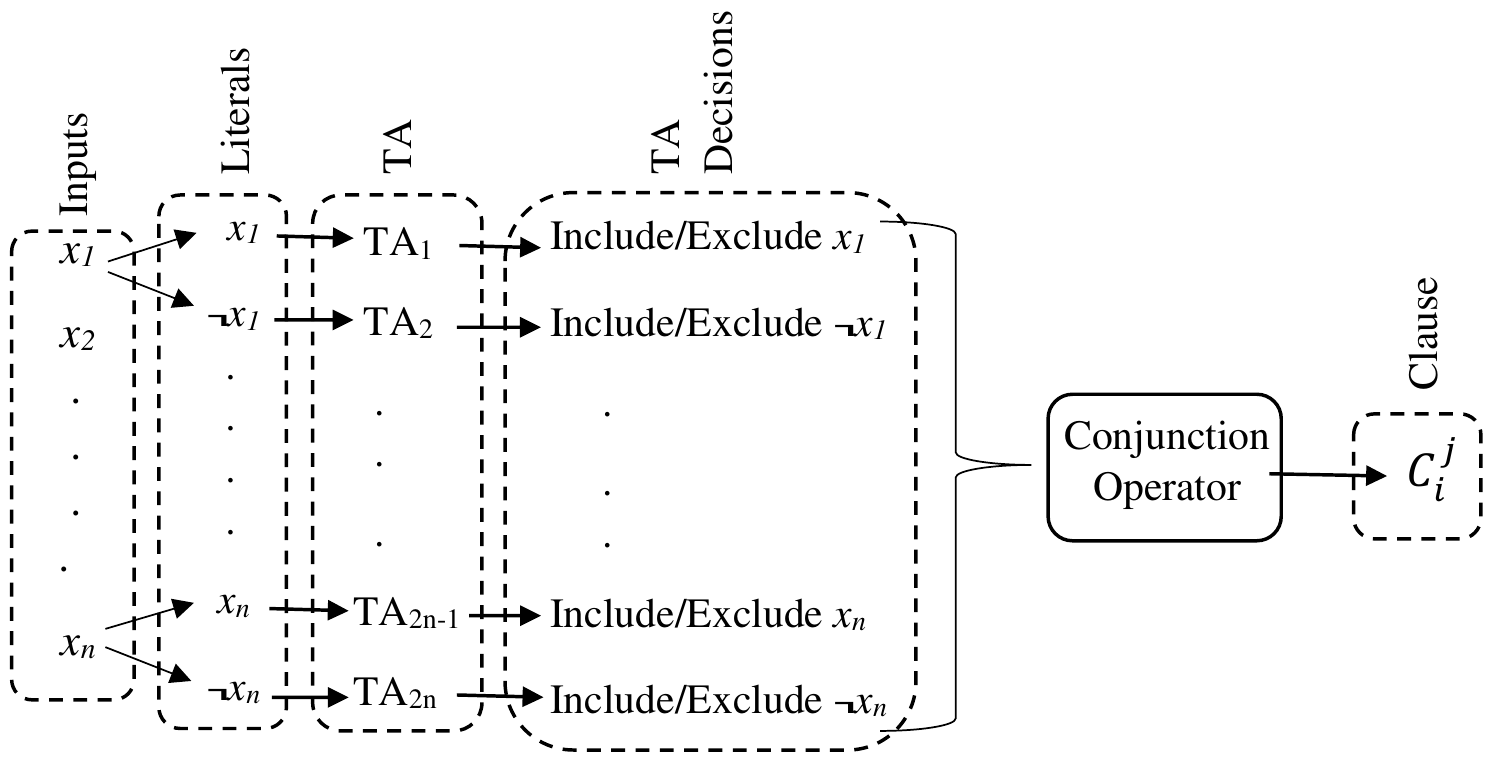}
\caption{}
\label{fig21}
\end{subfigure}\\[1ex]
\begin{subfigure}{.5\linewidth}
\centering
\includegraphics[scale=0.6]{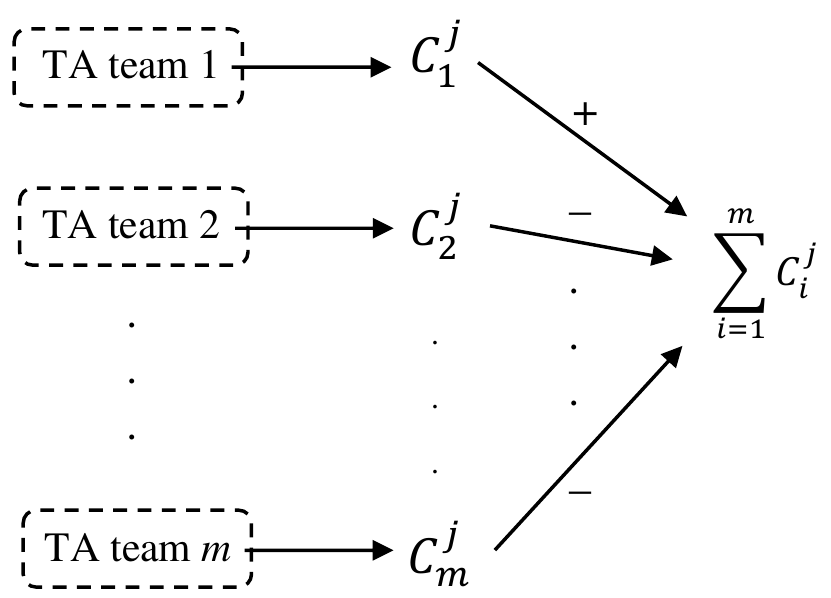}
\caption{}
\label{fig22}
\end{subfigure}%
\begin{subfigure}{.5\linewidth}
\centering
\includegraphics[scale=0.6]{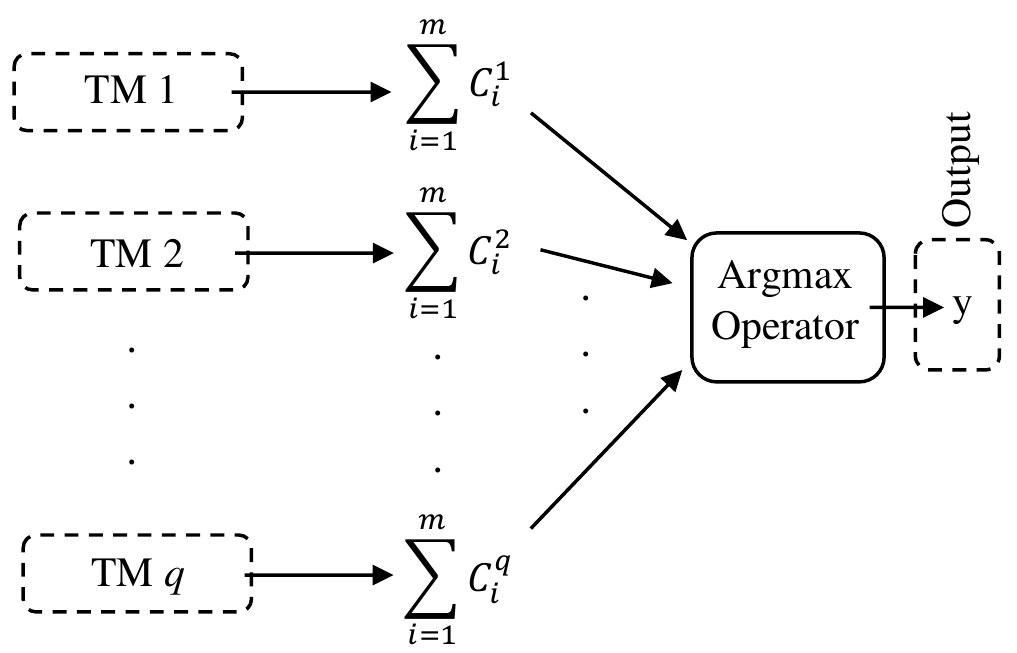}
\caption{}
\label{fig23}
\end{subfigure}
\vspace{-3mm}
\caption{(a) A TA team forms the clause $C_i^j$, 1 $\leq$ \textit{j} $\leq$ \textit{q}, 1 $\leq$ \textit{i} $\leq$ \textit{m}. (b) A TM. (c) A multiclass TM.}
\label{fig2}
\vspace{-5mm}
\end{figure}

In the TM, the disjunction operator is replaced by a summation operator to increase classification robustness \cite{Ole1}. The structure of the multiclass TM is depicted in Fig. \ref{fig22}. The sub-figures in Fig. \ref{fig2} illustrate the three phases of the classification process, i.e., (a) how a team of TAs forms a clause that processes the input features; (b) how a TM is composed by multiple TA teams; and (c) how a group of TMs are connected to handle multiclass classification problems. We will now detail these phases one by one.

\vspace{3mm}
\hspace{-5mm}\textbf{The TA team:}\\
\textit{Inputs and literals.} The TM takes $n$ propositional variables $x_1, x_2, x_3, \ldots, x_n$ as input. For each variable $x_k$, there are two literals, the variable itself and its negation $\lnot x_k$. 

\vspace{3mm}
\hspace{-6mm}\textit{Tsetlin Automata and their decisions.}
For each clause $C_i^j$, each literal is assigned a unique TA. This TA decides whether to \emph{include} or \emph{exclude} its assigned literal in the given clause. Thus, for $n$ input variables, we need $2n$ TA. This collective of TA is called a team. The TA team composes a conjunction of the literals that the team has chosen to be included. The conjunction outputs $1$ if all of the included literals evaluate to $1$, otherwise, the clause outputs $0$.

\vspace{3mm}
\hspace{-5mm}\textbf{The TM:}\\
\hspace{-6mm}\textit{Clauses and their role in a TM.} A TM consists of $m$ clauses, each associated with a TA team. The number of clauses needed for a particular class depends on the number of sub-patterns associated with the class. Each clause casts a vote, so that the $m$ clauses jointly decide the output of the TM. Clauses with odd indexes are assigned positive polarity $(+)$ and clauses with even indexes are assigned negative polarity $(-)$. The summation operator aggregates the votes by subtracting the number of negative votes from the number of positive votes. 

Note that clauses with positive polarity cast their votes to favor the decision that the input belongs to the class represented by the TM, whereas clauses with negative polarity vote for the input belonging to one of the other classes.

\vspace{3mm}
\hspace{-5mm}\textbf{The multiclass TM:}\\
\hspace{-6mm}\textit{Obtaining the final output.} With multiple TMs we get a multiclass TM. As shown in Fig. \ref{fig23}, the final decision is made by the argmax operator to classify the input data to the class that obtained the highest vote sum.

\subsection{The TA Game and Orchestration Scheme}

We organize learning in the TM as a game being played among the TAs. The Nash Equilibria of the game corresponds to the goal state of the TA, providing the final classifier. In the worst case, the single action of any TA has the power to disrupt the whole game. Therefore, the TAs must be guided carefully towards optimal pattern recognition.

To achieve this, the Tsetlin Machine is built around two kinds of feedback: Type I and Type II feedback. The Reward, Inaction, and Penalty probabilities under these two feedback types are summarized in Table \ref{Tab1}, and they are determined based on the clause output (1 or 0), the literal value (1 or 0), and the current action of the TA (include or exclude). Rewards and Penalties are fed to the TA as normal. Inaction means that the state of the TA remains unchanged.

The training process of the TM thus contains several interacting mechanisms. To clarify their roles, we provide a flow chart for the complete procedure, shown in Fig. 3, and explored in the following.
\begin{figure}[th!]
\centering
\includegraphics[width=11cm]{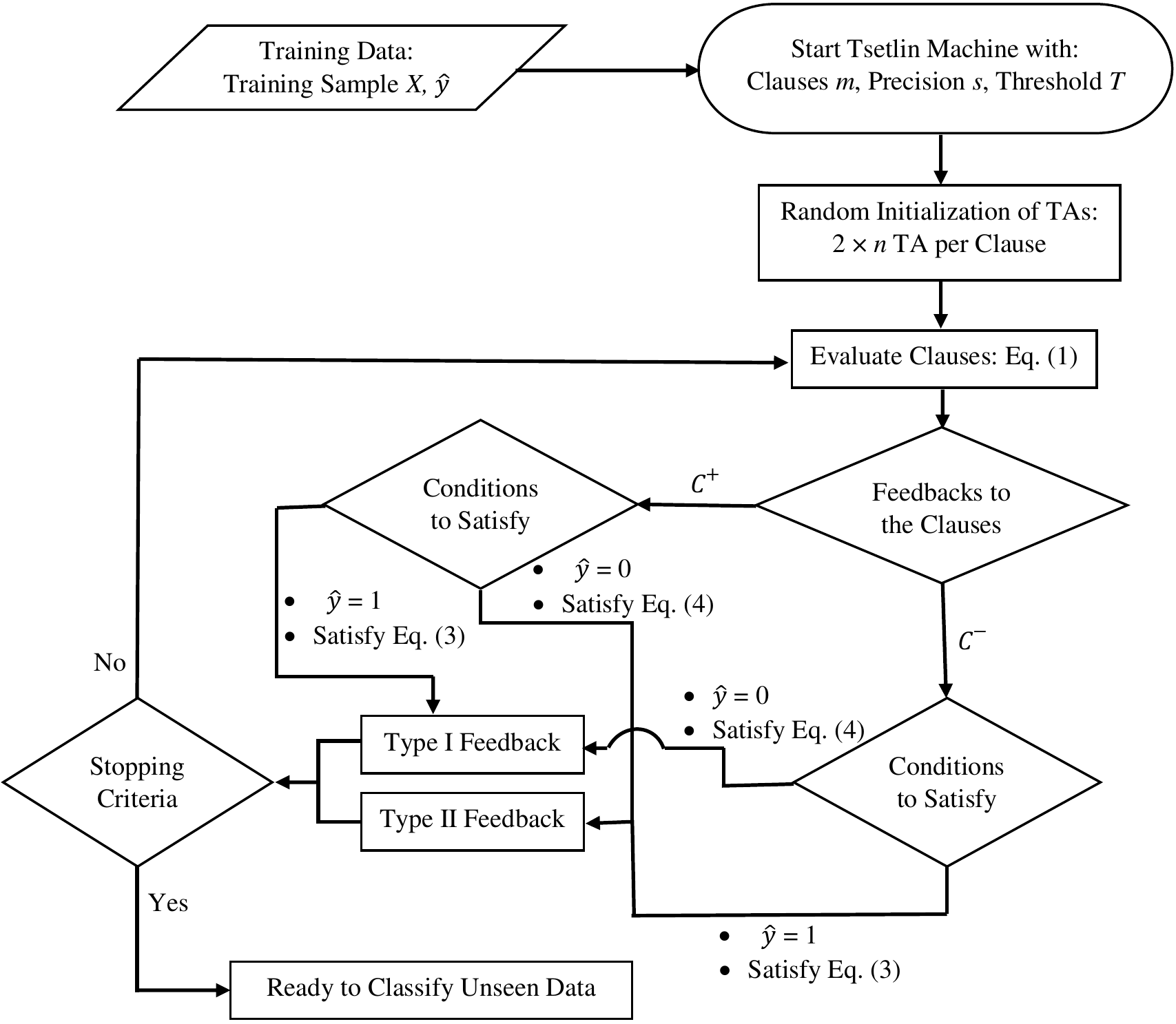}
\caption{The training work-flow.} \label{fig3}
\vspace{-5mm}
\end{figure}

{\bf Type I Feedback.} As seen in the flowchart, briefly stated, Type I feedback is only activated when the actual output $\hat{y}$ is 1. When the output of the target clause also is $1$, Type I Feedback has four roles:
\begin{itemize}
    \item It reinforces true positive output by assigning a large reward probability $\frac{s-1}{s}$ to the action of including literals that evaluate to $1$, and thus contributing to the result of the clause output being $1$.
    \item Conversely, exclude actions are penalized with the same magnitude under these conditions. This is to “tighten" the clause, because it would still output $1$ also when the literal considered is included instead.
    \item Furthermore, if the value of the literal is $0$, excluding the literal is the way to go, and exclude actions are thus rewarded with probability $\frac{1}{s}$.
\end{itemize}
When the output of the clause is $0$ (false negative output), Type I feedback has the following effect:
\begin{itemize}
    \item Type I feedback systematically penalizes include actions with probability $\frac{1}{s}$. Indeed, excluding literals is the only way to invert the output of a clause that outputs $0$.
    \item When the action is exclude, this is rewarded with probability $\frac{1}{s}$, because reinforcing exclude actions will sooner or later invert the output of the clause to $1$.
\end{itemize}
Thus, eventually, Type I feedback combats false negative output, and encourages true positive output.

{\bf Type II Feedback.} Type II feedback is activated when the actual output $\hat{y}$ is $0$, as shown in the flowchart. This type of feedback is designed to eliminate false positive output. That is, when the clause output should be $0$, but the clause erroneously evaluates to $1$, Type II feedback is triggered. In brief, repeated Type II feedback forces in the end the offending clause to evaluate to $0$, simply by including a literal that has the value $0$ into the clause (which makes the conjunction of literals evaluate to $0$ as well). This is achieved by penalizing, with probability $1$, exclude actions for literals that evaluate to $0$.

\begin{table}[t]
\centering
\newcolumntype{P}[1]{>{\centering\arraybackslash}p{#1}}
\caption{Type I and Type II feedback to battle against false negatives and false positives.}\label{Tab1}
\begin{tabular}{P{8mm}|c|c|c|c|c|c|P{6mm}|P{6mm}|P{6mm}|P{6mm}}
\toprule
\multicolumn{3}{c|}{Feedback Type} & \multicolumn{4}{c|}{I} & \multicolumn{4}{c}{II}  \\ \hline
\multicolumn{3}{c|}{Clause Output} & \multicolumn{2}{c|}{1} & \multicolumn{2}{c|}{0} & \multicolumn{2}{c|}{1} & \multicolumn{2}{c}{0} \\ \hline
\multicolumn{3}{c|}{Literal Value} & 1 & 0 & 1 & 0 & 1 & 0 & 1 & 0 \\ \hline
\multirow{6}{*}{\rotatebox[origin=c]{90}{Current State}} & \multirow{3}{*}{Include} & Reward Probability & (s-1)/s & NA & 0 & 0 & 0 & NA & 0 & 0 \\
                        &                     & Inaction Probability & 1/s & NA & (s-1)/s & (s-1)/s & 1 & NA & 1 & 1\\
                        &                     & Penalty Probability & 0 & NA & 1/s & 1/s & 0 & NA & 0 & 0 \\
                        & \multirow{3}{*}{Exclude} & Reward Probability & 0 & 1/s & 1/s & 1/s & 0 & 0 & 0 & 0\\
                        &                     & Inaction Probability & 1/s & (s-1)/s & (s-1)/s & (s-1)/s & 1 & 0 & 1 & 1\\
                        &                     & Penalty Probability & (s-1)/s & 0 & 0 & 0 & 0 & 1 & 0 & 0       \\ 
\bottomrule
\end{tabular}
\vspace{-5mm}
*s is the precision and controls the granularity of the sub-patterns in \cite{Ole1}
\end{table}

To summarize, Type I Feedback reinforces true positive output, while simultaneously reducing false negative output. These dynamics are countered by Type II Feedback, which systematically reduces false positive output.

{\bf The Clause Feedback Activation Function.} In \cite{Ole1}, an additional feedback mechanism is introduced, aiming at allocating the sparse pattern representation resources provided by the clauses as effectively as possible. This is achieved by introducing a target value $T$ for the number of clauses voting from a specific pattern. The idea is to gradually reduce the frequency of feedback for a specific pattern, as the number of votes approaches $T$. In all brevity, the feedback activation function is basically an activation probability controlled by a Threshold \textit{T}. The probability of activating Type I Feedback for a specific clause is:
\begin{equation}\label{eq3}
\frac{T - max(-T, min(T,\sum_{i=1}^{m} C_i^j))}{2T}
\vspace{2mm}
\end{equation}

For Type II Feedback, the probability is:

\begin{equation}\label{eq4}
\vspace{4mm}
\frac{T + max(-T, min(T,\sum_{i=1}^{m} C_i^j))}{2T}
\vspace{-2mm}
\end{equation}

As seen, for Eq. (\ref{eq3}), the activation probability decreases as the number of votes approaches \textit{T}, and finally when \textit{T} is reached, the probability becomes 0. Thus ultimately, Type I feedback will not be activated when enough clauses are producing the correct number of votes. This in turn “freezes" the affected clauses since TAs will no longer change state. The crucial point here is that this frees other clauses to seek other sub-patterns, because the “frozen" pattern is no longer attractive for the TA. The same rationale holds for Eq. (\ref{eq4}) for Type II feedback. In this way, the pattern representation resources can be allocated more effectively.

\subsection{Data Pre-Processing}

We now come to one of the main contributions of this paper, namely a scheme that allows the Tsetlin Machine to successfully recognize patterns consisting of continuous features, despite being constrained to an internal binary representation. As the TM only takes binary variables as input, we transform continuous features into binary form in a preprocessing step, detailed in the following.

Table \ref{Tab2} illustrates the transformation procedure, using one continuous feature as an example. The same procedure is repeated for each continuous feature in turn. First of all, all the unique values $\{v_1, v_2, \ldots, v_u\}$ of the continuous feature found in the dataset are identified. We consider each unique value $v_w$ to be a potential threshold “$\le  v_w$". Thus each unique value provides a new derived binary feature: is the threshold condition fulfilled or not fulfilled for a particular continuous value $v$.

As an example, column 1 in Table \ref{Tab2} contains the values of the continues features. As seen, there are three unique values, and these provides three thresholds $\le  3.834$, $\le 5.779$, and $\le  10.008$. Accordingly, three new binary features are introduced, encoding the original raw continuous values, also shown in the table. If the raw continuous value is greater than the threshold, the corresponding bit in the binary form is assigned the value $0$; and if the raw continuous value is less than or equal to the threshold, it is given the value  $1$. For example, in Table~\ref{Tab2}, for the first value $5.779$, it is greater than the first threshold $3.834$, so the corresponding binary feature is assigned the value $0$ (in column 2). However, being equal to the threshold value of the second threshold $5.779$, and less than the third threshold value $10.008$, both column 3 and column 4 are assigned the value $1$. Therefore, the final binary bits that represent 5.779 becomes $011$. Similarly, $10.008$ and $3.834$ are represented by $001$ and $111$, respectively.

This new representation becomes particularly powerful due to the capability of the TM to negate features, allowing a clause to specify intervals for continuous features. In the following section, we evaluate this procedure both on an artificial dataset, as well as for the real-life application of   forecasting dengue fever outbreaks in the Philippines.

\begin{table}[t]
\vspace{-8mm}
\centering
\newcolumntype{P}[1]{>{\centering\arraybackslash}p{#1}}
\caption{Conversion of original input features into bits.}\label{Tab2}
\begin{tabular}{P{15mm}|P{15mm}|P{15mm}|P{15mm}}
\toprule
\multirow{2}{*}{Raw Data} & \multicolumn{3}{c}{Thresholds}\\ \cline{2-4}
                        & $\le 3.834$ & $\le 5.779$ & $\le 10.008$ \\
\hline
5.779 & 0 & 1 & 1\\
10.008 & 0 & 0 & 1\\
5.779 & 0 & 1 & 1\\
3.834 & 1 & 1 & 1\\
\bottomrule
\end{tabular}
\vspace{-5mm}
\end{table}

\vspace{-3mm}
\section{Experiments}
\vspace{-1mm}

First, the behavior of the TM is studied using an artificial dataset. Actions chosen by TAs in clauses, clause outputs, and TM outputs, are extensively studied with this dataset. Then, the TM is applied to forecast the dengue outbreaks in the Philippines. Data and the TM preparation for both tasks are discussed in the following subsections.

\subsection{Behavior in Dealing with Artificial Data}

\subsubsection{Experimental Setup}\hfill \break

The dataset consists of two inputs (integers, 0 $\leq$ $x_1$ $\leq$ 4 and 0 $\leq$ $x_2$ $\leq$ 5). If the sum of the inputs is equal to 9, they are assigned class 1 and the rest is assigned class 0. Since the features in this process are categorical, we use one-hot-encoding to convert them into bits instead of the procedure proposed in the previous section. Input $x_1$ takes one of five values (0, 1, 2, 3, 4) and input $x_2$ takes one of six values (0, 1, 2, 3, 4, 5). Therefore, these two features can be expressed using 5 and 6 bits, respectively. An example data sample converted to bits can be found in Table \ref{Tab3}.

\begin{table}[b]
\vspace{-7mm}
\centering
\newcolumntype{P}[1]{>{\centering\arraybackslash}p{#1}}
\caption{Converting integer training samples to bits.}\label{Tab3}
\begin{tabular}{c|P{4mm}|P{4mm}|P{4mm}|P{4mm}|P{4mm}|P{4mm}|P{4mm}|P{4mm}|P{4mm}|P{4mm}|P{4mm}|c}
\toprule
\multirow{2}{*}{Original Sample} & \multicolumn{5}{|c}{$x_1$} & \multicolumn{6}{c|}{$x_2$} & $Out$\\ \cline{2-13}
                                & \multicolumn{5}{c|}{3} & \multicolumn{6}{c|}{5} & 8\\
\hline
Bit Positions & 0 & 1 & 2 & 3 & 4 & 0 & 1 & 2 & 3 & 4 & 5 & {}\\
Sample in Bits & 0 & 0 & 0 & 1 & 0 & 0 & 0 & 0 & 0 & 0 & 1 & 0\\
\bottomrule
\end{tabular}
\vspace{-3mm}
\end{table}

A Tsetlin Machine with 4 clauses is used to classify the artificial data. Since there are 11 input bits, 22 TAs are needed to form a clause. Each TA is given 100 states per action. Two of those four clauses will vote in favour of class 0. The other two clauses will vote in favour of class 1. The two remaining hyper parameters: Threshold and Precision are set to 1 and 8, respectively. 

\vspace{-3mm}
\subsubsection{Behavior Analysis}\hfill \break

\begin{figure}[t]
\centering
\includegraphics[width=9.5cm,height=7.5cm]{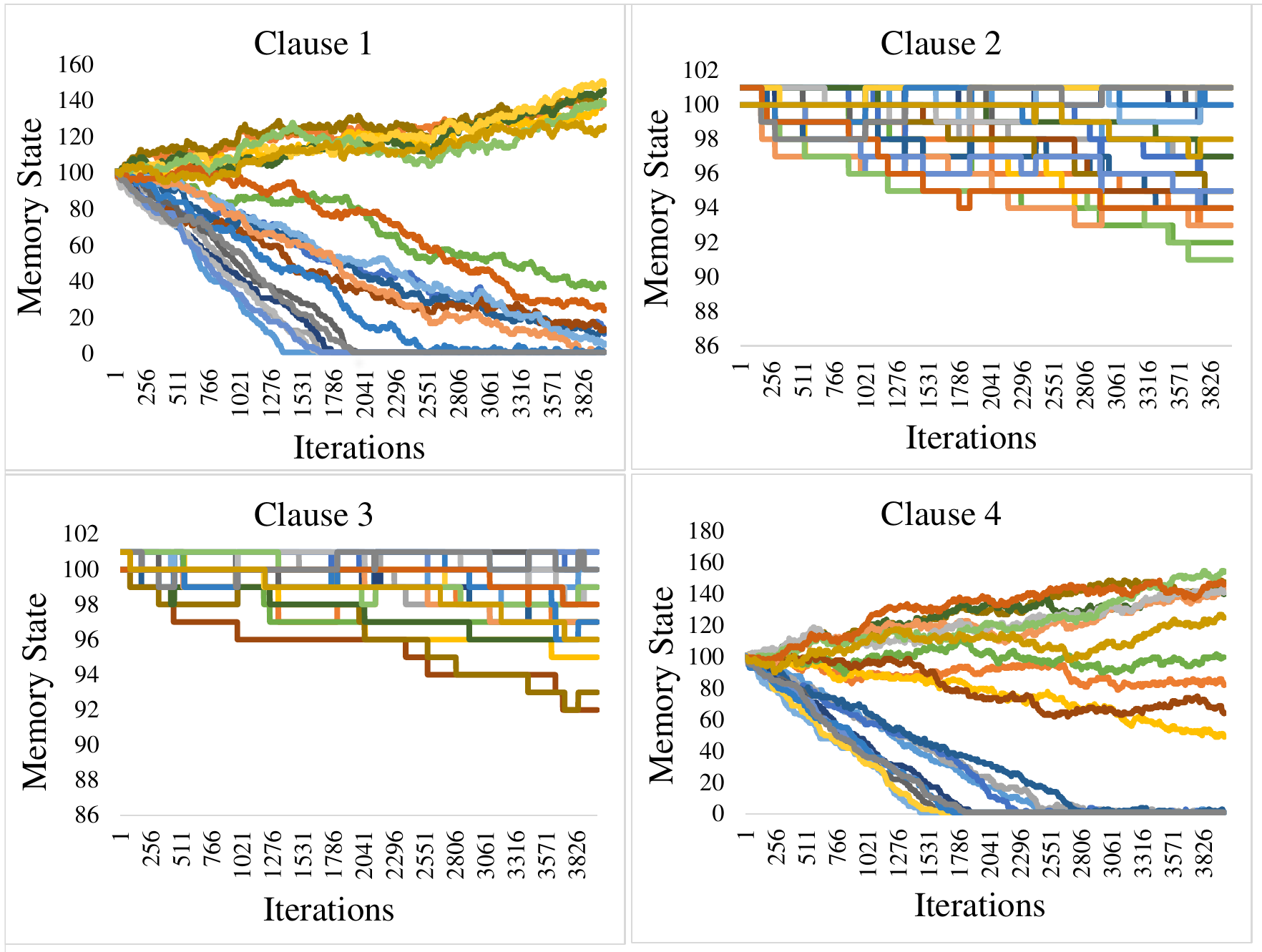}
\vspace{-3mm}
\caption{Variation of actions of TA to classify artificial data.} \label{fig4}
\vspace{-5mm}
\end{figure}

During the training process, the states of the TAs in each clause are recorded and plotted in Fig. \ref{fig4}. Clause 1 and 3 have positive polarities and clause 2 and 4 have negative polarities. Clause 1 and 4 vote in favour of class 0 while clause 2 and 3 vote in favour of class 1.

The change in states of the TAs in clause 1 and 4 are more dynamic compared to the TAs in clauses 2 and 3. This is due to the much larger number of training samples that belong to class 0. Since more training samples belong to class 0 ($\sim$ 8/9 of the data) than class 1 ($\sim$ 1/9 of the data), clauses 1 and 4 receive feedback more frequently. As seen, the TAs in clauses 2 and 3 move slowly towards the \textit{exclude} action (memory states from 0 to 100) since they receive Type II feedback more often (from class 0 data). However, once the TM is trained, it can classify all the 200 testing samples with 100\% accuracy.

\subsection{Predicting Disease Outbreaks}
\subsubsection{Experimental Setup}\hfill \break

The number of patients who suffer from dengue haemorrhagic fever or dengue shock syndrome has been increasing over the years. Therefore, dengue haemorrhagic fever is considered as an important public health issue, especially in tropical and subtropical countries. To control the mortality rate due to dengue fever, an early warning system, which helps directly on emergency preparedness and resource planning, is called for \cite{phung26}.

The Philippines has 17 administrative regions (from I to XVI with two IV regions; IVA and IVB). Department of Health in the Philippines has collected the number of monthly dengue incidences separately for all these regions from 2008 to 2016. 

The number of monthly dengue incidences in most of the regions has been growing from 2008 and peaked in 2013. However, from 2013, the number of incidences has dropped to reach an average value of 9.22 patients per 100,000 population. Considering these trends, we decided to use  more than 20 monthly dengue incidences per 100,000 population as an indication of outbreak. Using the data from 2008 to 2015, dengue outbreaks in the months of 2016 are to be predicted for all the regions. 

In addition to the dengue incidences in the previous-month and previous-year-same-month of the same region, historical dengue incidences from the neighboring regions and total dengue incidences are considered as input features to forecast the dengue outbreaks. These regions and total dengue incidences are selected based on their correlation to the target series. Dengue incidences in the previous-month of the selected regions and total dengue incidences are used as input features. The selected regions to forecast each region are summarized in Table \ref{Tab4}.

\begin{table}[!t]
\centering
\caption{Regions that provide their data to forecast dengue incidences of their neighbors}\label{Tab4} 
\begin{tabular}{c|c|c|c}	
\toprule
\textbf{Target} & \textbf{Selected Regions} & \textbf{Target} & \textbf{Selected Regions}\\ 
\hline
\textbf{I} & II,III, IVA, XIV, Total & \textbf{IX} & X, XI, XII\\

\textbf{II} & I, III, IVA, XIV, Total & \textbf{X} & IX, XI, XII, XIV, XV\\

\textbf{III} & I, II, IVA, XVI & \textbf{XI} & IX, X, XII, XV\\

\textbf{IVA} & III, IVB, V, XVI, Total & \textbf{XII} & IX, X, XI, XV, Total\\

\textbf{IVB} & II, IVA, VI, Total & \textbf{XIII} & IX, X, XII, XV, Total\\

\textbf{V} & IVA,VI, Total & \textbf{XIV} & I, II, III, IVA, IVB, Total\\

\textbf{VI} & IVB, V, VII, XII,Total & \textbf{XV} & IX, X, XI, XII\\

\textbf{XII} & IVA, V, VI, XII,Total & \textbf{XVI} & I, III, IVA, V\\

\textbf{VIII} & V, VI\\
\bottomrule

\end{tabular}
\vspace{-7mm}
\end{table}

Once the input features are determined, they are converted to bits using the procedure proposed in Section~3. Then they are fed into the TM to predicts dengue outbreaks in each region separately. Each TM has 2000 clauses and the
associated TAs are given 100 states per action. The other two hyper parameters, Threshold and Precision, are set to 15 and 8, respectively.

\vspace{-3mm}
\subsubsection{Results}\hfill \break

Possible dengue outbreaks in the Philippines for the year 2016 are forecasted by the TM. Results from the TM are compared with results from three other machine learning techniques: ANNs, a SVM, and a Decision Tree (DT). For comprehensiveness, four ANN architectures are used to forecast the dengue outbreaks: ANN-1 – one hidden layer with 5 neurons, ANN-2 – one hidden layer with 20 neurons, ANN-3 – three hidden layers with 20, 150, and 100 neurons, respectively, and ANN-4 – five hidden layers with 20, 200, 150, 100, and 50 neurons. The SVM uses a Radial Basis Function kernel to capture the non-linear patterns in the data. The regularization parameter (C) in this case is fixed at 1.0 with \emph{gamma} = 1/(the number of input features) to maximize prediction accuracy. The parameters which decide the quality of the DT output, such as maximum tree depth (=max), minimum number of samples required for split (=2), and the minimum number of samples required for a leaf node (=1) are again all adjusted to optimize prediction accuracy. Since there are 17 regions, all of the 204 testing samples are utilized to test the accuracy of each technique. These samples encompass 22 outbreaks to be identified. Each model is executed using 30-fold cross-validation to calculate precision, recall, F1-score, and accuracy. The means and the 95\% confidence intervals of these scores can be found in Table~\ref{Tab5}.

The TM obtains the highest mean values for precision and F1-score. The second highest precision (0.43) is obtained by the SVM, at the sacrifice of a much lower recall. Conversely, DT produces the highest recall, however, precision suffers. Considering overall performance, captured by the F1 score, the TM obtains the highest mean F1-score (0.40) while ANN-3 obtains the second highest mean F1-score (0.36). Even though mean F1 score peaks at 0.36, as a result of increasing the structural complexity of the ANNs, the score drops again when complexity is increased further. Due to the imbalance of the dataset (182 non-outbreaks and 22 outbreaks), the SVM produces a particularly high accuracy (0.89) by mostly classifying instances as non-outbreaks. Finally,  note that both the mean values of precision, recall, F1-score, and accuracy of the TM are higher than what we were able to achieve with the ANN models.

\begin{table}[t]
\vspace{-7mm}
\centering
\newcolumntype{P}[1]{>{\centering\arraybackslash}p{#1}}
\caption{Summary of the forecasting outcomes by different models.}\label{Tab5}
\begin{tabular}{c|c|c|c|c|c|c|c}
\toprule
 & TM &	ANN-1 &	ANN-2 &	ANN-3 &	ANN-4 &	SVM &	DT \\ \hline
 Precision & 0.44±0.02 &	0.35±0.02 &	0.39±0.01 &	0.37±0.02 &	0.36±0.02 &	0.43±0.01 &	0.29±0.02  \\
 Recall	& 0.37±0.02	& 0.23±0.02 &	0.31±0.02 &	0.36±0.02 & 0.33±0.03 &	0.14±0.01 &	0.41±0.02 \\
F1-score &	0.40±0.01 &	0.28±0.02 &	0.34±0.02 &	0.36±0.02 &	0.34±0.03 &	0.21±0.01 &	0.34±0.01 \\
Accuracy &	0.88±0.01 &	0.87±0.01 &	0.87±0.01 &	0.87±0.02 &	0.87±0.01 &	0.89±0.01 &	0.83±0.01 \\
\bottomrule
\end{tabular}
\vspace{-5mm}
\end{table}

\vspace{-2mm}
\section{Conclusion}
\vspace{-1mm}

In this paper, we proposed a feature pre-processing procedure for the TM so that it can effectively handle continuous input features. This opens up for promising applications in e.g. forecasting, where continuous features are typical. We applied the resulting TM approach to forecast dengue outbreaks in the Philippines, after performing an empirical study on an artificial dataset. While the experiments with the artificial dataset confirmed the desired properties of the new scheme, the results on the real-life dataset further demonstrated competitive performance also with respect to other machine learning approaches. Indeed, it turned out that the TM is more accurate than the evaluated SVMs, Decision Trees, and several multi-layered ANNs, both in terms of forecasting precision and F1-score.

In our further work, we intend to exploit this approach also in other pattern recognition domains where continuous features are dominant. We further intend to investigate how also the output of the TM can be rendered continuous.


\vspace{-4mm}

\begin{thebibliography}{}

\bibitem{Ole1} O.-C. Granmo, “The Tsetlin Machine - A Game Theoretic Bandit Driven Approach to Optimal Pattern Recognition with Propositional Logic,” \textit{arXiv:1804.01508}.\

\bibitem{narendra3} K.  S.  Narendra  and  M.  A.  Thathachar, ``Learning  Automata:  An  Introduction''. \textit{Courier Corporation}, 2012.\

\bibitem{Berge2018}G.  Thore  Berge,  O.-C.  Granmo,  T.  Oddbjørn  Tveit,  M.  Goodwin,  L.  Jiao,  and B. Viggo Matheussen, ``Using the Tsetlin Machine to Learn Human-Interpretable Rules for High-Accuracy Text Categorization with Medical Applications'', \textit{arXiv:1809.04547}, Sep 2018.\

\bibitem{ogihara7}H. Ogihara, Y. Fujita, Y. Hamamoto, N. Iizuka, and M. Oka, ``Classification Based on Boolean Algebra and Its Application to the Prediction of Recurrence of Liver Cancer'',  \textit{2013  2nd  IAPR  Asian  Conference  on Pattern  Recognition  (ACPR)}, pp. 838–841, IEEE, 2013.\

\bibitem{cruz8}R. S. Cruz, B. Fernando, A. Cherian, and S. Gould, “Neural Algebra of Classifiers,” \textit{arXiv:1801.08676}, 2018.\

\bibitem{wang6}T.  Wang,  C.  Rudin,  F.  Doshi-Velez,  Y.  Liu,  E.  Klampfl,  and  P.  MacNeille,  “A Bayesian Framework for Learning Rule Sets for Interpretable Classification,” \textit{The Journal of Machine Learning Research}, vol. 18, no. 1, pp. 2357–2393, 2017.\

\bibitem{feldman9}V. Feldman, “Hardness of Approximate Two-Level Logic Minimization and PAC Learning with Membership Queries,” \textit{Journal  of  Computer  and  System  Sciences}, vol. 75, no. 1, pp. 13–26, 2009.\

\bibitem{klivans10}A. R. Klivans and R. A. Servedio, “Learning DNF in Time 2o (n1/3),” \textit{Journal of Computer and System Sciences}, vol. 68, no. 2, pp. 303–318, 2004.\

\bibitem{feldman11} V. Feldman, “Learning DNF Expressions From Fourier Spectrum,” in \textit{Conference on Learning Theory}, pp. 17–1, 2012.\

\bibitem{valiant12} L. G. Valiant, “A Theory of the Learnable,” \textit{Communications of the ACM}, vol. 27,no. 11, pp. 1134–1142, 1984.\

\bibitem{hauser13}J. R. Hauser, O. Toubia, T. Evgeniou, R. Befurt, and D. Dzyabura, “Disjunctions of Conjunctions, Cognitive Simplicity, and Consideration Sets,” \textit{Journal of Marketing Research}, vol. 47, no. 3, pp. 485–496, 2010.\

\bibitem{rudin14}C. Rudin, B. Letham, and D. Madigan, “Learning Theory Analysis for Association  Rules  and  Sequential  Event  Prediction,” \textit{The  Journal  of  Machine  Learning Research}, vol. 14, no. 1, pp. 3441–3492, 2013.\

\bibitem{mccormick15} T. McCormick, C. Rudin, and D. Madigan, “A Hierarchical Model for Association Rule Mining of Sequential Events: An Approach to Automated Medical Symptom Prediction,” \textit{Annals of Applied Statistics}, 2011.\

\bibitem{granmo16}O.-C. Granmo and B. J. Oommen, “Solving Stochastic Nonlinear Resource Allocation Problems Using a Hierarchy of Twofold Resource Allocation Automata.,” \textit{IEEE Transaction on Computers}, 2010.\

\bibitem{oommen19}B. J. Oommen, S.-W. Kim, M. T. Samuel, and O.-C. Granmo, “A Solution to the Stochastic  Point  Location  Problem  in  Metalevel  Nonstationary  Environments,” \textit{IEEE  Transactions  on  Systems,  Man,  and  Cybernetics,  Part  B  (Cybernetics)},vol. 38, no. 2, pp. 466–476, 2008.\

\bibitem{tung17} B.  Tung  and  L.  Kleinrock,  “Using  Finite  State  Automata  to  Produce  Self-Optimization and Self-Control,” \textit{IEEE transactions on parallel and distributed systems}, vol. 7, no. 4, pp. 439–448, 1996.\

\bibitem{bouhmala18}N. Bouhmala and O.-C. Granmo, “Stochastic Learning for SAT-Encoded Graph Coloring  Problems,” \textit{International  Journal  of  Applied  Metaheuristic  Computing(IJAMC)}, vol. 1, no. 3, pp. 1–19, 2010.\

\bibitem{darshana20}K.  Abeyrathna,  O.-C.  Granmo,  and  M.  Goodwin,  “A  novel  Tsetlin  Automata Scheme to Forecast Dengue Outbreaks in the Philippines,” 2018 IEEE 30th Inter-national Conference on Tools with Artificial Intelligence (ICTAI), 2018.\

\bibitem{gharbi21}M.  Gharbi,  P.  Quenel,  J.  Gustave,  S.  Cassadou,  G.  La  Ruche,  L.  Girdary,  andL. Marrama, “Time Series Analysis of Dengue Incidence in Guadeloupe, French West  Indies:  Forecasting  Models  Using  Climate  Variables  as  Predictors,” \textit{BMC infectious diseases}, vol. 11, no. 1, p. 166, 2011.\

\bibitem{choudhury22} Z. M. Choudhury, S. Banu, and A. M. Islam, “Forecasting Dengue Incidence in Dhaka, Bangladesh: A Time Series Analysis.,” \textit{Dengue Bulletin}, vol. 32, 2008.\

\bibitem{luz23}P. M. Luz, B. V. Mendes, C. T. Code co, C. J. Struchiner, and A. P. Galvani, “Time Series  Analysis  of  Dengue  Incidence  in  Rio  de  Janeiro,  Brazil,” \textit{The  American journal of tropical medicine and hygiene}, vol. 79, no. 6, pp. 933–939, 2008.\

\bibitem{silawan24} T.  Silawan,  P.  Singhasivanon,  J.  Kaewkungwal,  S.  Nimmanitya,  and  W.  Su-wonkerd,  “Temporal  Patterns  and  Forecast  of  Dengue  Infection  in  Northeastern Thailand,” \textit{Southeast  Asian  Journal  of  Tropical  Medicine  and  Public  Health},vol. 39, no. 1, p. 90, 2008.\

\bibitem{promprou25} S. Promprou, M. Jaroensutasinee, and K. Jaroensutasinee, “Forecasting Dengue Hemorrhagic Fever Cases in Southern Thailand Using ARIMA Models.,” \textit{Dengue Bulletin}, vol. 30, 2006.\

\bibitem{phung26} D. Phung, C. Huang, S. Rutherford, C. Chu, X. Wang, M. Nguyen, N. H. Nguyen,and C. Do Manh, “Identification of the Prediction Model for Dengue Incidence in Can Tho City, a Mekong Delta Area in Vietnam,” \textit{Acta tropica}, vol. 141, pp. 88–96,2015.\

\bibitem{abeyrathna27}K.  Abeyrathna,  O.-C.  Granmo,  and  M.  Goodwin,  “Effect  of  Data  From  Neighbouring Regions to Forecast Dengue Incidences in Different Regions of Philippines Using Artificial Neural Networks,” \textit{2018: Norsk Informatikkonferanse}, 2018.\


\end{thebibliography}
\end{document}